\documentclass[runningheads]{llncs}
\usepackage{graphicx}
\usepackage{amsmath}
\usepackage{amssymb}
\begin{document}
\title{Deep Iterative 2D/3D Registration}
\titlerunning{Deep Iterative 2D/3D Registration}

\author{Srikrishna~Jaganathan \inst{1,2} \and Jian~Wang\inst{2} \and Anja~Borsdorf\inst{2} \and Karthik~Shetty\inst{1} \and Andreas~Maier\inst{1}}


\authorrunning{S. Jaganathan et al.}

\institute{Pattern~Recognition~Lab, FAU~Erlangen-Nürnberg, Erlangen, Germany.\\
\email{srikrishna.jaganathan@fau.de}
\and Siemens~Healthineers~AG, Forchheim, Germany. \\
}

\maketitle       
\begin{abstract}

Deep Learning-based 2D/3D registration methods are highly robust but often lack the necessary registration accuracy for clinical application. A refinement step using the classical optimization-based 2D/3D registration method applied in combination with Deep Learning-based techniques can provide the required accuracy. However, it also increases the runtime. In this work, we propose a novel Deep Learning driven 2D/3D registration framework that can be used end-to-end for iterative registration tasks without relying on any further refinement step. We accomplish this by learning the update step of the 2D/3D registration framework using Point-to-Plane Correspondences. 
The update step is learned using iterative residual refinement-based optical flow estimation, in combination with the Point-to-Plane correspondence solver embedded as a known operator. Our proposed method achieves an average runtime of around 8s, a mean re-projection distance error of $0.60 \pm 0.40$~mm with a success ratio of 97 percent and a capture range of 60~mm. The combination of high registration accuracy, high robustness, and fast runtime makes our solution ideal for clinical applications. 

\end{abstract}

\section{Introduction}

In X-ray-based image-guided interventions, 2D \mbox{X-ray} fluoroscopy is preferable for providing image guidance. However, due to the projective nature of the 2D images acquired, there is an inherent loss of information. Overlaying the preoperative 3D volume onto the 2D images can provide the necessary additional information during the intervention. To obtain an accurate overlay, one needs to register the preoperative 3D volume with the current patient position, which is accomplished using 2D/3D registration. The goal of 2D/3D registration is to find the optimal transformation of the preoperative 3D volume to the current patient position.

The problem of 2D/3D registration for medical images has already been explored for decades and has well-established techniques for specific use cases~\cite{markelj2012review}. In many of these classical techniques, the registration problem is often formulated as an optimization problem and solved with iterative schemes. If the initial misalignment is large, the optimization problem is often non-convex, thus requiring global optimizers to avoid the problem of getting stuck in local minima~\cite{markelj2012review}. Such global optimizers are computationally expensive and are not sufficiently fast for application during the intervention if the inital registration error is large.

Deep Learning (DL) based techniques for 2D/3D registration problem have shown promising results, by improving the computational efficiency~\cite{miao2016real,schaffert2020learning} and robustness~\cite{miao2018dilated,liao2019multiview,corrs_schaffert}. Recent works have also shown that the robustness can be increased to a much greater extent using DL-based methods and even propose fully automatic 2D/3D registration solution~\cite{esteban2019towards,grimm2020pose}.  However, most of these techniques rely on a final refinement step based on the classical methods to achieve the necessary registration accuracy for interventional application, which limits the computational efficiency of DL-based registration.

The recent trend has clearly shown that DL-based techniques can provide good initialization and, when used in combination with a refinement step, can provide a hybrid 2D/3D registration method.
A DL-based registration method that doesn't rely on refinement for registration accuracy is highly desirable. We propose a DL-based solution to accomplish this goal, where we learn the update step of the iterative 2D/3D registration framework based on Point-to-Plane Correspondence (PPC) constraint~\cite{wang2017dynamic}. Learning such an update step prediction, proposed in~\cite{jaganathan2021learning}, showed significant improvement (around two times) for single-step update prediction. However, the iterative application using the learned update step to the actual 2D/3D registration problem was lacking. We retain the structure of the previously proposed update step prediction~\cite{jaganathan2021learning} but make significant architectural changes to incorporate another domain-prior, which approximately models the iterative nature of the problem. For this, we use iterative residual refinement~\cite{hur2019iterative} based optical flow estimation. In combination with the PPC solver embedded as a known operator, we can learn the update step end-to-end directly on the registration loss.

\section{Methods}

\begin{figure}
    \centering
    \includegraphics[width=\linewidth]{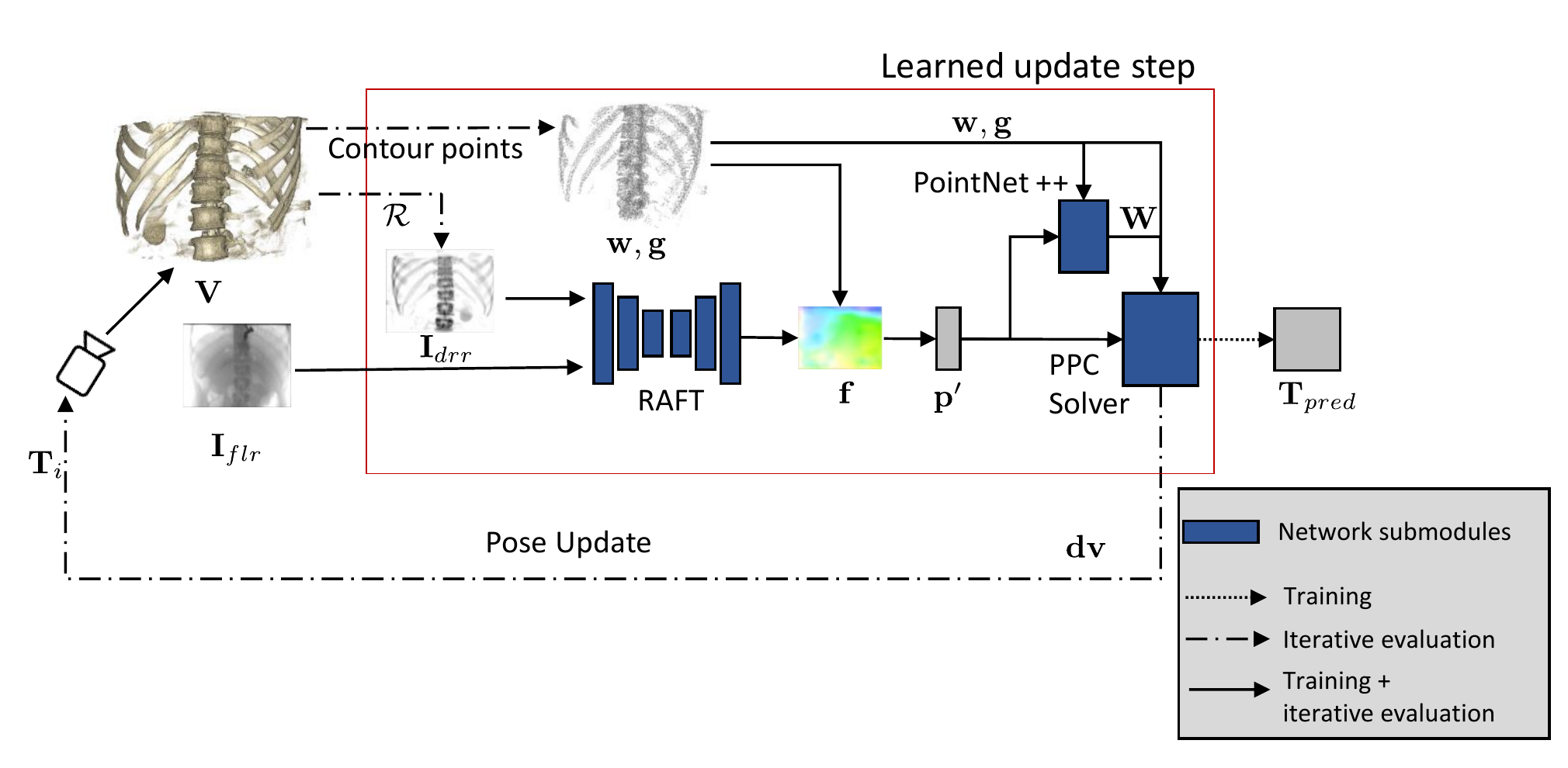}
    \caption{Overview of the proposed registration framework with the DL-driven learned update step prediction ${\phi}((\mathbf{I}_{flr},\mathbf{I}_{drr},\mathbf{w},\mathbf{g}), \mathbf{\mathcal{W}})$.
    The RAFT architecture $\mathbf{\phi}_f$ is used for correspondence estimation and the PointNet++ architecture $\mathbf{\phi}_w$ is used to estimate per correspondence weight matrix $\mathbf{W}$. The differentiable PPC solver $\mathcal{K}_{ppc}$ computes the 3D motion update.}
    \label{fig:registration_framework}
\end{figure}

We use the PPC-based iterative registration framework~\cite{wang2017dynamic} along with our \mbox{DL-based} update step prediction. The schematic of the proposed registration framework is depicted in Fig.~\ref{fig:registration_framework}. The proposed \mbox{DL-based} solution acts as a \mbox{drop-in} replacement for the update step prediction in the original framework. We provide a brief background on the original framework~\cite{wang2017dynamic}, what constitutes an update step, and the PPC constraint in the following section. 

\subsection{Background}

 The input to the framework is the fluoroscopic \mbox{X-ray} image $\mathbf{I}_{\mathrm{flr}}$, the preoperative CT volume $\mathbf{V}$ along with a rough initial registration estimate $\mathbf{T}_{\mathrm{init}}$. We assume the intrinsic camera parameters $\mathbf{K}$ is known.
 The framework consists of an initialization step and an update step. During the initialization, the surface points along with its gradients are extracted from $\mathbf{V}$. Digitally Reconstructed Radiograph (DRR) $\mathbf{I}_{\mathrm{drr}} =\mathcal{R}(\mathbf{V},\mathbf{T}_i)$ is rendered from $\mathbf{V}$ and the current pose $\mathbf{T}_i$ at iteration $i$ using the rendering operator $\mathcal{R}$. Pose dependent apparent contour points $\mathbf{w},$ with the corresponding gradients $\mathbf{g}$ are selected from the surface points~\cite{wang2017dynamic} before the start of each update step.
 
The update step $\mathcal{U}(\mathbf{I}_{flr},\mathbf{I}_{drr},\mathbf{w},\mathbf{g})$ predicts the 3D motion update $\mathbf{dv}$ given ($\mathbf{I}_{flr},\mathbf{I}_{drr},\mathbf{w},\mathbf{g}$) as inputs to the update step. The update step has the following operations. Initially, 2D correspondences search is performed between $\mathbf{I}_{\mathrm{flr}}$ and $\mathbf{I}_{\mathrm{drr}}$ at the projected contour points $\mathbf{p} =\mathbf{\mathcal{P}} (\mathbf{w},\mathbf{K})$ in $\mathbf{I}_{\mathrm{drr}}$, where $\mathcal{P}$ is the projection operation. This results in a set of correspondences ($\mathbf{p},\mathbf{p}'$), where $\mathbf{p}'$ is the corresponding points of $\mathbf{p}$ found in $\mathbf{I}_{\mathrm{flr}}$. The 2D correspondences can only reveal the observable 2D misalignment. However, to recover the true 3D motion error and thus finding the optimal transformation, the unobservable 3D motion components should also be considered. The total 3D motion is effectively constrained using the PPC model. The PPC model constraints the 3D motion to a plane spanning between $ (\mathbf{w} \times \mathbf{g}) $ and $\mathbf{p}'$~\cite{wang2017dynamic}. We use the weighted variant of the PPC constraint to reduce the effects of noisy 2D correspondences. The weighted PPC constraint is as follows:

\begin{equation}
    \mathbf{W}~\mathbf{A} \mathbf{dv} =  \mathrm{diag}(\mathbf{W})~b\enspace,
    \label{equ:ppc_constraint}
\end{equation}
where $\mathbf{A} = ((\mathbf{n} \times \mathbf{w}^\intercal) - \mathbf{n}^\intercal)$ and $\mathbf{b}=\mathbf{n}^T \mathbf{w}$. The matrix $\mathbf{A} \in {\bbbr}^{N \times 6}$ and vector $\mathbf{b} \in {\bbbr}^{N}$ are computed from the $N_\mathrm{cp}$ point correspondences ($\mathbf{p},\mathbf{p}'$), $\mathbf{n}$ is the plane normal. The weight matrix $\mathbf{W}$ is a diagonal matrix providing individual weights for each estimated correspondence. The 3D motion vector $\mathbf{dv}$ can be computed from the PPC constraint using \mbox{closed-form} solution with \mbox{pseudo-inverse} of $\mathbf{A}$ and converted to a transformation matrix $\mathbf{T}_{i+1}$ which generates the input to the next update step. The process is repeated until convergence criteria is reached~\cite{wang2017dynamic}.

\subsection{DL-Based Update Step Prediction}

We parameterize $\mathcal{U}$ with ${\phi}(\mathcal{U}, \mathbf{\mathcal{W}})$, where $\phi$ is the Deep Neural Network (DNN) and $\mathbf{\mathcal{W}}$ are the network parameters. \mbox{DNN-based} parameterization for $\mathcal{U}$ was proposed earlier~\cite{jaganathan2021learning}, however we have significant changes in the network architecture to model the iterative nature of the problem more precisely. The network $\phi$ is composed of two sub networks, $\phi_f$ for correspondence estimation, $\phi_w$ for correspondence weighting and a PPC solver layer $\mathcal{K}_{ppc}$ which has no learnable weights. We describe the different submodules in the following paragraphs.

\subsubsection{Correspondence Estimation Network $\mathcal{\phi}_f$}

We use the recently proposed Recurrent All-Pairs Field Transforms (RAFT)~\cite{teed2020raft} architecture for optical flow estimation between $\mathbf{I}_{\mathrm{drr}}$ and $\mathbf{I}_{\mathrm{flr}}$. The network uses iterative residual refinement~\cite{hur2019iterative} for estimating the optical flow. The network consists of a recurrent component (referred as update operator in~\cite{teed2020raft}) which unrolls the flow prediction and enables the use of residual refinement.
The flow prediction is iteratively refined $\mathbf{f}_{\mathrm{j+1}} = \mathbf{f}_j + \Delta \mathbf{f} $ by repeated application of the update operator until it converges to a fixed point (mimicking the behavior of optimization methods). At each update, the network only needs to predict the small residual $\Delta \mathbf{f}$. During training, the update operator is unrolled for a fixed number of iterations $N_{\mathrm{FL}}$. We use a version of the RAFT architecture with shared weights between the updates, as it is both memory efficient and gives the best performance~\cite{teed2020raft}. Since pixel level ground truth flow is not available, we use sparse ground truth flow labels at the projected contour points $\mathbf{p}$ which is directly computed from the ground truth registration matrix $\mathbf{T}_\mathrm{gt}$. We use a modified optical flow loss function compared to~\cite{teed2020raft} to counter the sparsity of the labels~\cite{corrs_schaffert}.
The network outputs dense correspondence for all pixels, but we sample the predicted flow $\mathbf{f}$ at the last flow update step $N_{\mathrm{FL}}$ of the network, to compute flow $\mathbf{dp}$ at the projected contour points $\mathbf{p}$. We find the corresponding points $\mathbf{p}' = \mathbf{p} + \mathbf{dp}$ in $\mathbf{I}_{flr}$ and use it as input to the next layers.

\subsubsection{Correspondence Weighing $\mathcal{\phi}_w$}
Our correspondence weighting network is based on the PointNet++ architecture~\cite{qi2017pointnet++}, which takes in as input feature vector $\mathbf{f}_\mathrm{{\phi_w}}= \{\mathbf{w},\mathbf{g},\mathbf{nw},\mathbf{p}'\} \in \bbbr^{N_{\mathrm{cp}}\times9}$.
The PointNet++ architecture performs hierarchical feature learning on 3D point cloud data and incorporates local context information, which was missing in the original PointNet~\cite{qi2017pointnet}. It can handle additional features along with the 3D Euclidean coordinate features thus we don't need any modification for our input feature vector. We use single scale grouping variant of the PointNet++ segmentation architecture, which can already provide per point label. We only modify the final activation to sigmoid activation to predict correspondence weights. The network predicts per correspondence weight similar to the attention model~\cite{schaffert2020learning}. The weight matrix $\mathbf{W}$ obtained from a set of $N_{\mathrm{cp}}$ contour points is used to solve the weighted PPC model in Eq.~\eqref{equ:ppc_constraint}.

\subsubsection{PPC solver $\mathcal{K}_{ppc}$}
The differentiable PPC layer takes in the correspondence set $\{\mathbf{p},\mathbf{p}'\}$, the contour points with its gradients $\{\mathbf{w},\mathbf{g}\}$ and the estimated per correspondence weights $\mathbf{W}$ and solves for $\mathbf{dv}$ from the PPC constraint (Eq.~\eqref{equ:ppc_constraint}) using \mbox{closed-form} solution. The motion vector $\mathbf{dv}$ is converted to a transformation matrix $\mathbf{T}_{pred}$ which serves as the predicted output from our proposed network $\mathcal{\phi}$.

\subsection{Loss Function}

The registration loss is computed between the contour point positions at $\mathbf{T}_\mathrm{gt}$ and $\mathbf{T}_\mathrm{pred}$ using $\mathcal{L}_{reg}=|| \mathbf{T}_{\mathrm{pred}} (\mathbf{w}) -  \mathbf{T}_{\mathrm{gt}} (\mathbf{w})||_1$.
We use a modified average End Point Error (EPE) to compute optical flow (Eq.~\eqref{equ:flow_loss}) loss at the projected contour points. A binary mask image $ \mathbf{M}_{\mathrm{\mathbf{p}}}$ of the projected contour points is used to zero out the loss at other pixels~\cite{corrs_schaffert}. Since the RAFT architecture has an unrolled update operator, which produces flow at $N_{\mathrm{FL}}$ as $\{\mathbf{f}_j\}$, we use the same ground truth flow for all iteration with a discount factor $\gamma^{N_{\mathrm{FL}} - j}$~\cite{teed2020raft}. The optical loss is computed as follows,

\begin{equation}
    \mathcal{L}_{flow} = \sum_{j=1}^{N_{\mathrm{FL}}} \gamma^{N_{\mathrm{FL}}-j}  \frac{1}{N_{\mathrm{cp}}}  \mathbf{M}_{\mathrm{\mathbf{p}}} {|| \mathbf{f}^j - \mathbf{f}_{gt} ||}_{1} \enspace. 
    \label{equ:flow_loss}
\end{equation}

The combined training loss function with the regularizers is as follows, 
\begin{equation}
    \mathcal{L} = \alpha~\mathcal{L}_{\mathrm{flow}} + \beta~|| \mathbf{T}_{\mathrm{pred}} (\mathbf{w}) -  \mathbf{T}_{\mathrm{gt}} (\mathbf{w})||_1 + \lambda~||\mathbf{dv}||_2 + \frac{\zeta}{2}~||\mathbf{\mathcal{W}}||^2  \enspace,
    \label{equ:combined_loss}
\end{equation}
where we use a regularization on the estimated $\mathbf{dv}$ computed from the PPC solver and weight decay regularizer on the network weights $\mathbf{\mathcal{W}}$.  The hyper-parameters $\alpha$, $\beta$ are used to control the strength of optical flow and registration loss respectively. The hyper-parameters $\lambda$, $\zeta$ are used to control the  motion and weight decay regularizer strength respectively.

\section{Experiments and Results}

We validate our method using clinical cone beam CT (CBCT) reconstruction data set for a single view scenario, since it is harder for the registration methods and largely remains unsolved compared to multi-view scenario which was the focus of many of the previous works~\cite{miao2018dilated,liao2019multiview,schaffert2019robust}.

\subsection{Data}

 The data set consists of reconstructed CBCT volumes and the corresponding \mbox{X-ray} images used for reconstruction. The ground truth registration between the \mbox{X-ray} images and the CBCT volume is available. The data set is from vertebra body region, consisting of 55 patient volumes (includes both thoracic and lumbar regions). The number of \mbox{X-ray} images per CBCT volume varies depending on the slice thickness, between 190 to 390. The slice resolution also varies between $256 \times 256$ to $512 \times 512$ and the voxel spacing  between $0.49$~mm  to $0.99$~mm for all the dimension. The \mbox{X-ray} images has a resolution of $616 \times 480$ with a pixel spacing of $0.616$~mm. Random initial transformation is used to create training samples with initial registration error measured in mean Target Registration Error (mTRE)~\cite{van2004standardized} is in the range of $[0,30]$~mm, with translations in range of $[0, \pm 30]$~mm and rotation in range of $[0,\pm 20]$ degrees for all the three axes. Each training sample consists of the initial transformation matrix $\mathbf{T}_{\mathrm{init}}$, the ground truth registration matrix $\mathbf{T}_\mathrm{gt}$,($ \mathbf{w},\mathbf{g}$), $\mathbf{I}_\mathrm{drr}$ rendered based on $\mathbf{T}_{\mathrm{init}}$ and $\mathbf{I}_\mathrm{flr}$. From each patient we generate around 1200 to 1800 such samples using the random initial transformations depending on the number of fluoroscopic images available. The data set is split into training, validation and test sets with 43 patients for training, 6 patients for validation, 6 patients for testing. We have combined total of around 80,000 samples for training and validation.

\subsection{Training}

 We pretrain $\mathbf{\phi}_f$ for 50 epochs on our training data using Eq.~\eqref{equ:flow_loss}. This is to avoid the \mbox{pseudo-inverse} computation used in PPC solver from failing when the correspondence estimation is bad, which can be the case when $\phi_f$ is randomly initialized. Following this we train our proposed network $\mathcal{\phi}$ with the loss function described in Eq.~\eqref{equ:combined_loss} for 100 epochs with early stopping criteria used on validation data set. The hyper-parameters of the loss function (Eq.~\eqref{equ:combined_loss}) used are set to $\alpha = 1$, $\beta = 0.5$, $\lambda=1e-3$ and $\zeta = 1e-5$ and $\gamma$ in Eq.~\eqref{equ:flow_loss} set to 0.8~\cite{teed2020raft}. We unroll $\mathbf{\phi}_f$ with $N_{\mathrm{FL}}=6$ iterations for both training and evaluation. ADAM~\cite{kingma2014adam} optimizer is used with a cyclical learning rate varying from $1e-4$ to $1e-6$ and a batch size of 16. We implemented the network using the PyTorch framework~\cite{paszke2019pytorch}. We compute the gradients of all the layers by back-propagation directly using autograd~\cite{paszke2017automatic} in PyTorch.

Data augmentation plays a crucial role for us to overcome the limited amount of training samples available considering the number of network parameters and the complex nature of the problem. Online data augmentation is used with color space transforms (where we adjust brightness and contrast of both $I_{drr}$ and $I_{flr}$), geometric transforms (affine 2D rotation with translation, horizontal and vertical flips) and random erasing~\cite{zhong2020random}. The corresponding modification in ground truth data is performed to account for the augmentation.

\subsection{Evaluation}
\label{subsec:eval}

We use the standardized evaluation measures~\cite{van2004standardized} for 2D/3D registration. The initial error range is reported in mTRE and the final registration error is measured in mean Re-Projection Distance (mRPD)~\cite{van2004standardized} as it is standard practice for the evaluation of single view registration ~\cite{van2004standardized,corrs_schaffert,schaffert2020learning}. Along with the registration error, we also indicate the Success Ratio (SR) and the Capture Range (CR) to quantify the robustness.
We define the success threshold as mRPD $\leq$ 2.0~mm final registration error~\cite{schaffert2020learning}. The capture range measures the highest sub-interval of initial error for which we can achieve SR $\geq$ 95\%. Due to the large initial error range considered here, we only report capture range at the intervals of 5mm. We run all the evaluations using the Intel Core i7-6850K CPU and Nvidia GeForce Gtx Titan X GPU with 12 GB graphics memory and report the average run time for the registration. Our proposed network is evaluated for iterative 2D/3D registration on the test data set (6 Patients). For each patient, 600 test samples are created from ground truth registration using random initial transformations of the AP and LAT views. The initial registration error measured in mTRE varies from [0,60]~mm with translations in the range of $[0, \pm 60]$~mm and rotation in the range of $[0,\pm 40]$ degrees for all three axes. We run our proposed method for 10 iterations.

We compare our proposed method with the existing state-of-the-art techniques. The considered models, as well as their method used to compute the update step, are described below. We consider DPPC~\cite{wang2020robust}, the classical \mbox{depth-aware} PPC model which relies on patch matching for correspondence estimation and heuristics for correspondence weighting, DPPC Attention~\cite{schaffert2020learning} uses patch matching for correspondence estimation and PointNet based learned correspondence weighting, DPPC CL Refinement (DPPC-CL-A in~\cite{corrs_schaffert}) which uses a FlowNet~\cite{dosovitskiy2015flownet} architecture to estimate the correspondence, however, a refinement step is required using patch-matching and heuristics are used for correspondence weighting~\cite{corrs_schaffert}. PPC Flow Attention~\cite{jaganathan2021learning} which uses FlowNetC~\cite{dosovitskiy2015flownet} architecture for correspondence estimation and PointNet based attention~\cite{schaffert2020learning} for correspondence weighting. The baseline methods were run for the proposed evaluation configurations in the respective work. We use pretrained weights obtained from the authors of the respective baseline methods, trained on the same data set for the learnable modules in baseline methods. In this way, we ensure that the most optimized version of the baseline methods are used.

\subsection{Results}

\begin{table}[!h]
\centering
\begin{tabular}{p{0.33\linewidth}p{0.18\linewidth}p{0.12\linewidth}p{0.14\linewidth}p{0.18\linewidth}}
\hline 
  & mRPD [mm]~$\downarrow$ & SR [\%] ~$\uparrow$  & CR [mm]~$\uparrow$  & Runtime [s]~$\downarrow$  \\
 & $\displaystyle \mu \ \pm \sigma $ &  &   & $\displaystyle \mu \ \pm \sigma $ \\
\hline
 DPPC~\cite{wang2020robust} & $0.58 \pm 0.218$  & 61.9 & 10-15 & $36.13 \pm 26.20 $ \\
 DPPC CL Refinement~\cite{corrs_schaffert}& $\mathbf{0.40 \pm 0.09}$  & 94.58 & 20-25 & $37.53 \pm 11.40$ \\
 DPPC Attention~\cite{schaffert2020learning}& $0.47 \pm 0.23$ & 95.5 & 35-40  & $16.42 \pm 23.43$ \\
PPC Flow Attention~\cite{jaganathan2021learning}& $1.60 \pm 0.34$ & 35.4 & 0 & $12.82\pm 0.2 $ \\
Proposed method & $0.60 \pm 0.40$  & $\mathbf{97.0}$ & $\mathbf{55-60}$ & $ \mathbf{8.05\pm 0.2}$  \\
\hline
\\
\end{tabular}
\caption{Comparison of our proposed method with the existing state-of-the-art techniques. The initial error range measured in mTRE varies between $[0,60]$~mm. We report final registration error in mRPD~[mm], Success Ratio (SR) (mRPD $\leq$ 2.0~mm), Capture Range (CR) and the average runtime~[s] for solving one registration problem. $\uparrow$ indicates that higher values are better and $\downarrow$ indicates that lower values are better.}
\label{table:results_comparisons}
 
\end{table}

\begin{figure}[!h]
    \centering
    \includegraphics[width=\linewidth]{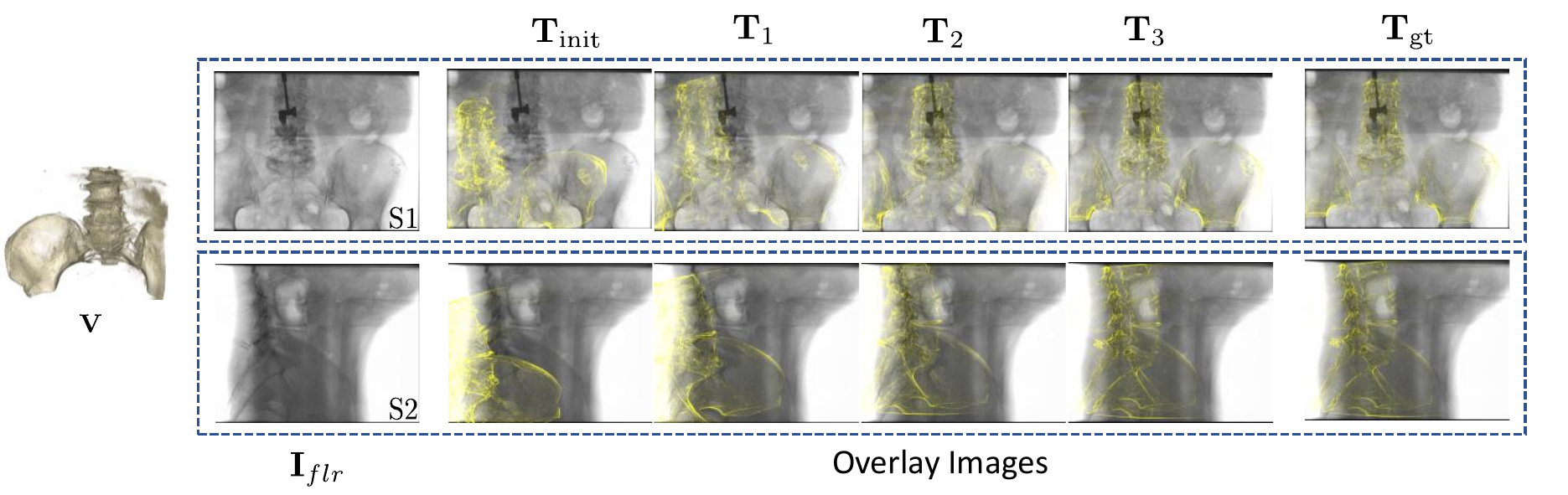}
    \caption{Overview of the qualitative results shown on two example cases S1 (AP) and S2 (LAT) on a test data set patient using our proposed technique. $\mathbf{T}_\mathrm{init}$ shows a large inital misalignment for both the examples. The ground truth overlay is also provided for comparison. We show the first three iterations which demonstrates the speed of convergence from large initial registration error.}
    \label{fig:example_img}
\end{figure}

We evaluate our proposed methods and the baseline methods as described in Section~\ref{subsec:eval}.
The evaluation results of all the considered methods are presented in Table~\ref{table:results_comparisons}. DPPC CL Refinement~\cite{corrs_schaffert} achieves the best registration accuracy of $0.47 \pm 0.23$~mm. Our proposed methods has the best SR with 97\%, CR with 55-60~mm and average runtime for a registration problem at $8.05\pm 0.2$~s. Qualitative results of our proposed registration method is shown in Fig.~\ref{fig:example_img}, evaluated for two different views on one test data set patient.

\section{Discussion and Conclusion}

The results presented in Table~\ref{table:results_comparisons} show that our proposed method improves the SR by 1.5 \% and increases the capture range by 20~mm (50\% improvement) compared to the other state-of-the-art methods without sacrificing registration accuracy, as we achieve sub-millimeter registration error. Our technique is robust for a comparatively higher range of initial error. We are also twice faster compared to other methods, which is a crucial factor for the interventional application. Our proposed technique performs significantly better, compared to \cite{jaganathan2021learning} which fails for the iterative registration task using a similar learned update step. This shows modeling the iterative nature of the problem is essential for such a learned update step, as this allows the network to learn intermediate flow steps, thus allowing the network to perform well for both large and small displacements.

In summary, we proposed a DL-driven iterative 2D/3D registration framework which is fast, robust and provides highly accurate registration. To the best of our knowledge, we are one of the first DL-driven method to retain high robustness and also achieve highly accurate registration without any further refinement.
Future research direction can be extending the proposed method to fully automatic registration and multi-view scenario. One challenge for using the proposed method, is that, it requires a large number of annotated training data. Use of simulated data in combination with domain randomization~\cite{grimm2020pose} or adversarial data augmentation~\cite{peng2018jointly} strategies can be explored to reduce the burden of annotated training data requirements.

\subsubsection{Disclaimer:} The concepts and information presented in this paper are based
on research and are not commercially available.

%
%
%
\bibliographystyle{splncs04}
\bibliography{paper2357}

\begin{thebibliography}{10}
\providecommand{\url}[1]{\texttt{#1}}
\providecommand{\urlprefix}{URL }
\providecommand{\doi}[1]{https://doi.org/#1}

\bibitem{dosovitskiy2015flownet}
Dosovitskiy, A., Fischer, P., Ilg, E., Hausser, P., Hazirbas, C., Golkov, V.,
  Van Der~Smagt, P., Cremers, D., Brox, T.: Flownet: Learning optical flow with
  convolutional networks. In: Proceedings of the IEEE international conference
  on computer vision. pp. 2758--2766 (2015)

\bibitem{esteban2019towards}
Esteban, J., Grimm, M., Unberath, M., Zahnd, G., Navab, N.: Towards fully
  automatic x-ray to ct registration. In: International Conference on Medical
  Image Computing and Computer-Assisted Intervention. pp. 631--639. Springer
  (2019)

\bibitem{grimm2020pose}
Grimm, M., Esteban, J., Unberath, M., Navab, N.: Pose-dependent weights and
  domain randomization for fully automatic x-ray to ct registration. arXiv
  preprint arXiv:2011.07294  (2020)

\bibitem{hur2019iterative}
Hur, J., Roth, S.: Iterative residual refinement for joint optical flow and
  occlusion estimation. In: Proceedings of the IEEE/CVF Conference on Computer
  Vision and Pattern Recognition. pp. 5754--5763 (2019)

\bibitem{jaganathan2021learning}
Jaganathan, S., Wang, J., Borsdorf, A., Maier, A.: Learning the update operator
  for 2d/3d image registration. In: Bildverarbeitung f{\"u}r die Medizin 2021.
  pp. 117--122. Springer Fachmedien Wiesbaden, Wiesbaden (2021)

\bibitem{kingma2014adam}
Kingma, D.P., Ba, J.: Adam: A method for stochastic optimization. arXiv
  preprint arXiv:1412.6980  (2014)

\bibitem{van2004standardized}
Van~de Kraats, E.B., Penney, G.P., Toma{\v{z}}evi{\v{c}}, D., van Walsum, T.,
  Niessen, W.J.: Standardized evaluation of 2d-3d registration. In:
  International Conference on Medical Image Computing and Computer-Assisted
  Intervention. pp. 574--581. Springer (2004)

\bibitem{liao2019multiview}
Liao, H., Lin, W.A., Zhang, J., Zhang, J., Luo, J., Zhou, S.K.: Multiview 2d/3d
  rigid registration via a point-of-interest network for tracking and
  triangulation. In: Proceedings of the IEEE/CVF Conference on Computer Vision
  and Pattern Recognition. pp. 12638--12647 (2019)

\bibitem{markelj2012review}
Markelj, P., Toma{\v{z}}evi{\v{c}}, D., Likar, B., Pernu{\v{s}}, F.: A review
  of 3d/2d registration methods for image-guided interventions. Medical image
  analysis  \textbf{16}(3),  642--661 (2012)

\bibitem{miao2018dilated}
Miao, S., Piat, S., Fischer, P., Tuysuzoglu, A., Mewes, P., Mansi, T., Liao,
  R.: Dilated fcn for multi-agent 2d/3d medical image registration. In:
  Proceedings of the AAAI Conference on Artificial Intelligence. vol.~32 (2018)

\bibitem{miao2016real}
Miao, S., Wang, Z.J., Zheng, Y., Liao, R.: Real-time 2d/3d registration via cnn
  regression. In: 2016 IEEE 13th International Symposium on Biomedical Imaging
  (ISBI). pp. 1430--1434. IEEE (2016)

\bibitem{paszke2017automatic}
Paszke, A., Gross, S., Chintala, S., Chanan, G., Yang, E., Devito, Z., Lin, Z.,
  Desmaison, A., Antiga, L., Lerer, A.: Automatic differentiation in pytorch
  (2017)

\bibitem{paszke2019pytorch}
Paszke, A., Gross, S., Massa, F., Lerer, A., Bradbury, J., Chanan, G., Killeen,
  T., Lin, Z., Gimelshein, N., Antiga, L., et~al.: Pytorch: An imperative
  style, high-performance deep learning library. arXiv preprint
  arXiv:1912.01703  (2019)

\bibitem{peng2018jointly}
Peng, X., Tang, Z., Yang, F., Feris, R.S., Metaxas, D.: Jointly optimize data
  augmentation and network training: Adversarial data augmentation in human
  pose estimation. In: Proceedings of the IEEE Conference on Computer Vision
  and Pattern Recognition. pp. 2226--2234 (2018)

\bibitem{qi2017pointnet}
Qi, C.R., Su, H., Mo, K., Guibas, L.J.: Pointnet: Deep learning on point sets
  for 3d classification and segmentation. In: Proceedings of the IEEE
  conference on computer vision and pattern recognition. pp. 652--660 (2017)

\bibitem{qi2017pointnet++}
Qi, C.R., Yi, L., Su, H., Guibas, L.J.: Pointnet++: Deep hierarchical feature
  learning on point sets in a metric space. arXiv preprint arXiv:1706.02413
  (2017)

\bibitem{schaffert2020learning}
Schaffert, R., Wang, J., Fischer, P., Borsdorf, A., Maier, A.: Learning an
  attention model for robust 2-d/3-d registration using point-to-plane
  correspondences. IEEE transactions on medical imaging  \textbf{39}(10),
  3159--3174 (2020)

\bibitem{schaffert2019robust}
Schaffert, R., Wang, J., Fischer, P., Maier, A., Borsdorf, A.: Robust
  multi-view 2-d/3-d registration using point-to-plane correspondence model.
  IEEE transactions on medical imaging  \textbf{39}(1),  161--174 (2019)

\bibitem{corrs_schaffert}
Schaffert, R., Wei{\ss}, M., Wang, J., Borsdorf, A., Maier, A.: Learning-based
  correspondence estimation for 2-d/3-d registration. In: Bildverarbeitung
  f{\"u}r die Medizin 2020, pp. 222--228. Springer (2020)

\bibitem{teed2020raft}
Teed, Z., Deng, J.: Raft: Recurrent all-pairs field transforms for optical
  flow. In: European Conference on Computer Vision. pp. 402--419. Springer
  (2020)

\bibitem{wang2020robust}
Wang, J.: Robust 2-D/3-D Registration for Real-time Patient Motion
  Compensation: Robuste 2-D/3-D Registrierung Zur Echtzeitf{\"a}higen,
  Dynamischen Bewegungskompensation. Verlag Dr. Hut (2020)

\bibitem{wang2017dynamic}
Wang, J., Schaffert, R., Borsdorf, A., Heigl, B., Huang, X., Hornegger, J.,
  Maier, A.: Dynamic 2-d/3-d rigid registration framework using point-to-plane
  correspondence model. IEEE transactions on medical imaging  \textbf{36}(9),
  1939--1954 (2017)

\bibitem{zhong2020random}
Zhong, Z., Zheng, L., Kang, G., Li, S., Yang, Y.: Random erasing data
  augmentation. In: Proceedings of the AAAI Conference on Artificial
  Intelligence. vol.~34, pp. 13001--13008 (2020)

\end{thebibliography}

\end{document}